\documentclass[letterpaper, 10 pt, conference]{ieeeconf}  

\IEEEoverridecommandlockouts                              
\usepackage{bm}
\usepackage{latexsym,amssymb,amsmath}
\usepackage{arydshln}
\usepackage{threeparttable}
\usepackage[ruled]{algorithm2e}
\usepackage[noend]{algpseudocode}
\usepackage{lipsum,multicol}
\usepackage{graphicx}
\usepackage{multirow}
\usepackage{balance}

\usepackage{amsmath} 
\usepackage{hyperref} 
\usepackage{mathtools} 
\usepackage{booktabs} 
\DeclareMathOperator*{\argmin}{arg\,min} 

\usepackage{xcolor}
\definecolor{myBlue}{rgb}{0 0.4470 0.7410}
\definecolor{myRed}{rgb}{0.6350 0.0780 0.1840}

\usepackage{marvosym}

\makeatletter
\let\MYcaption\@makecaption
\makeatother

\usepackage[font=footnotesize]{subcaption}

\makeatletter
\let\@makecaption\MYcaption
\makeatother

\title{\LARGE \bf
Increasing SLAM Pose Accuracy by Ground-to-Satellite Image Registration
}

\author{Yanhao Zhang$^{1} \textsuperscript{(\Letter)}$, Yujiao Shi$^{2}$, Shan Wang$^{3,4}$, Ankit Vora$^{5}$, \\
	Akhil Perincherry$^{5}$, Yongbo Chen$^{3}$, and Hongdong Li$^{3}$
\thanks{
$^{1}$Yanhao Zhang is with the Robotics Institute, University of Technology Sydney, Sydney, Australia (e-mail: yanhao.zhang@uts.edu.au).
}
\thanks{
$^{2}$ Yujiao Shi is with ShanghaiTech University, Shanghai, China (e-mail: shiyj2@shanghaitech.edu.cn). 
}
\thanks{
$^{3}$ Shan Wang, Yongbo Chen, and Hongdong Li are with the College of Engineering and Computer Science, Australian National University, Canberra, Australia (e-mails: shan.wang@anu.edu.au; yongbo.chen@anu.edu.au; hongdong.li@anu.edu.au).}
\thanks{$^{4}$Shan Wang is also with Data61, CSIRO, Canberra, Australia.}
\thanks{$^{5}$Ankit Vora and Akhil Perincherry are with Ford Motor Company, Dearborn, USA (e-mails: avora3@ford.com; aperinch@ford.com).}
\thanks{This work was performed while Yanhao Zhang and Yujiao Shi worked at the Australian National University.
}
}%

\begin{document}

\maketitle
\thispagestyle{empty}
\pagestyle{empty}

\begin{abstract}
Vision-based localization for autonomous driving has been of great interest among researchers. When a pre-built 3D map is not available, the techniques of visual simultaneous localization and mapping (SLAM) are typically adopted. Due to error accumulation, visual SLAM (vSLAM) usually suffers from long-term drift. This paper proposes a framework to increase the localization accuracy by fusing the vSLAM with a deep-learning based ground-to-satellite (G2S) image registration method. In this framework, a coarse (spatial correlation bound check) to fine (visual odometry consistency check) method is designed to select the valid G2S prediction. The selected prediction is then fused with the SLAM measurement by solving a scaled pose graph problem. To further increase the localization accuracy, we provide an iterative trajectory fusion pipeline. The proposed framework is evaluated on two well-known autonomous driving datasets, and the results demonstrate the accuracy and robustness in terms of vehicle localization.
The code will be available at \url{https://github.com/YanhaoZhang/SLAM-G2S-Fusion}.
%
%
%
\end{abstract}


\section{Introduction} \label{sec_intro}
Accurate localization is an essential task for autonomous driving. Although GPS is widely used in people's daily lives, the accuracy of a consumer-grade GPS device can degrade rapidly in GPS-compromised areas \cite{xiong2021g}, e.g., the urban areas with high-rising buildings, which does not meet the requirements for autonomous driving \cite{reid2019localization}. Alternatively, other techniques adopts a pre-rendered 3D high-definition (HD) map for vehicle re-localization \cite{wolcott2014visual, pascoe2015direct, liu2017efficient, liu2023cyberloc}. However, it is laborious and expensive to reconstruct and maintain such an HD map, and a pre-built map only supports the vehicle's re-localization. Therefore, the study on self-localization techniques using only on-board sensors for autonomous driving is of great interest among researchers. 


Simultaneous localization and mapping (SLAM) is one of the major topics in robotics for sensor self-localization. Among different sensors, cameras are usually cheaper while capturing richer information. The task of visual SLAM (vSLAM) is to build and update a 3D map while simultaneously estimating the camera's trajectory, using the 2D features extracted from the consecutive input images. Since SLAM results are based on the consecutive observation of the same scene, without any global information, the observation error accumulates over time resulting in long-term drift. This is especially a problem for autonomous driving when a vehicle moves from one place to another without any loops.

\begin{figure}[t]  
\centering
\includegraphics[width=1.0\linewidth]{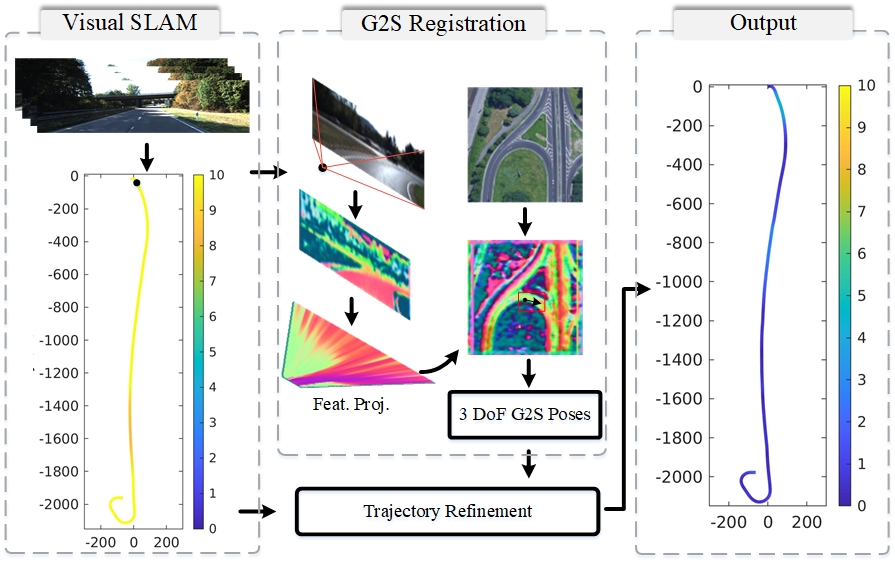}
\caption{The proposed framework fuses vSLAM with G2S registration and estimates the camera trajectory with high accuracy. 
The inputs are poses from stereo SLAM, ground-view images, and satellite images, the output is an updated vehicle trajectory. The example shows the localization error using the colour map (unit: m).}
\label{fig01_framework}
\end{figure}


Recently, the problem of ground-to-satellite (G2S) registration has aroused attention in academics. It aims to estimate the 3-DoF pose of a ground-view image w.r.t. a satellite image. 
Compared to the localization methods relying on an HD map, the cross-view based solution utilizes relatively cheaper and more memory-efficient satellite images. 
The satellite images can also provide global information for the camera pose reference, making it a preferable option to eliminate the long-term drift from SLAM estimation.



\begin{figure*}[t]
	\centering
 \setlength{\abovecaptionskip}{0pt}
    \setlength{\belowcaptionskip}{0pt}
	\includegraphics[width=1.0\linewidth]{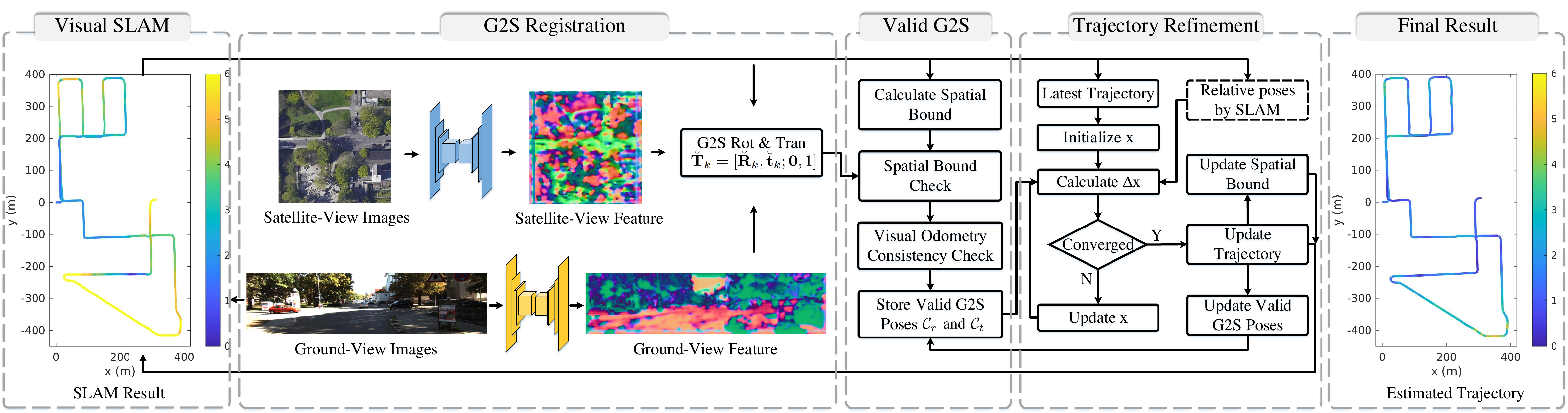} 
	\caption{A flowchart showing the main processes of the proposed framework. For each vehicle pose, we calculate the G2S prediction $\breve{\mathbf{T}}$ using the ground-view and the corresponding satellite images. The valid predictions are selected via a coarse-to-fine procedure, and are fused with the relative poses (from the original SLAM trajectory) by solving a scaled pose graph optimization. The localization error is shown using the colour map (unit: m).}
	\label{fig02_method}
\end{figure*}

This paper proposes a novel framework for camera pose estimation that combines the merits of vSLAM and G2S registration (Fig. \ref{fig01_framework}). 
vSLAM experiences long-term drift issues while achieving acceptable accuracy in the short term. In contrast, G2S registration minimizes error accumulation (it concentrates on pose estimation between two views), while the predicted translation, especially along the longitudinal axis, is not robust enough as shown in \cite{shi2023boosting}. Our purpose is to fuse the two methods together and provide a more accurate localization framework for autonomous driving. 
Particularly, we want to use the pose estimated by G2S registration to rectify the pose estimated by SLAM. 

Considering the pose estimated by G2S registration is not always more accurate than SLAM, we design a coarse-to-fine mechanism to select the valid G2S poses, which is different from the existing methods \cite{hu2020image, sarlin2023orienternet}. 
Specifically, we first coarsely select the G2S poses based on a spatial bound determined by the trajectory uncertainty, which are then further refined by checking the visual odometry consistency. 
After this coarse-to-fine filtering, the selected G2S poses are considered valid and fused with the SLAM poses by solving a scaled pose graph problem. 
%
To increase the localization accuracy, we update the trajectory iteratively when a G2S pose is considered valid. 
The proposed framework is evaluated on two well-known autonomous driving datasets, and the results demonstrate that our method achieves around 68\%-80\% improvement in translation estimation and 45\%-65\% on rotation estimation compared to the original vSLAM.  

The main contributions of this paper are as follows:
\begin{itemize}
	\item a new localization framework that combines the merits of vSLAM and G2S image registration;
	\item a coarse-to-fine method to remove false G2S results;
	\item an iterative trajectory refinement pipeline that fuses the measurements by solving a scaled pose graph problem.
\end{itemize}

\section{Related Work}

\subsection{Visual Localization}
There has been intensive research on visual localization methods for autonomous driving. Among these methods, vSLAM has traditionally been studied, e.g., \cite{campos2021orb, engel2014lsd, leutenegger2015keyframe, mur2015orb, mur2017orb, qin2017vins, mur2017visual, wang2023mavis}, where ORB-SLAM3 \cite{campos2021orb} is the SOTA vSLAM system consisting of visual tracking, local mapping, and loop closure. Deep learning based localization has also been investigated to handle large appearance changes between the query images and the reference maps \cite{von2020gn, von2020lm, sarlin2021back}. The main idea is to train a neural network and use the implicitly calculated correspondence (by the deep features between a query image and a reference image) for camera pose estimation. Similar ideas are also shown to be effective for G2S registration \cite{shi2022beyond, wang2023satellite}.

\subsection{G2S-based Visual Localization}
The early works on G2S registration solve an image retrieval problem and achieve a coarse localization \cite{hu2018cvm, liu2019lending, shi2019spatial, toker2021coming, shi2020optimal}. Later, fine-grained methods are proposed to estimate the relative pose from a coarse input pose. To calculate the G2S correspondences, some methods use an additional sensor, e.g., LiDAR, \cite{wang2023satellite, fervers2022continuous}, while others vision-only methods calculate feature matches using homography transformations \cite{shi2022beyond, xia2022visual, sarlin2023orienternet} or sparse view-consistent keypoints from a pre-trained network \cite{wang2023view}. Recent research has shown that decoupling the estimated rotation and translation can facilitate the overall registration performance \cite{shi2023boosting, fervers2023uncertainty}. BoostG2SLoc \cite{shi2023boosting} is the SOTA pipeline where a neural pose optimizer is deployed to estimate the azimuth orientation, and a spatial correlation module is used to predict the longitudinal and lateral translation.

\section{Problem Description}

Given a coarse pose between a paired ground-view and satellite images $\{I^g, I^s\}$, a deep neural network is adopted to predict the relative transformation between $I^g$ and $I^s$.
To be more specific, the output of the G2S prediction is the changes of rotation and translation w.r.t. the input pose. 
Let $\tilde{\mathbf{T}}_{k} = [\tilde{\mathbf{R}}_{k}, \tilde{\mathbf{t}}_{k}; \mathbf{0}, 1]$ denote a (input) SLAM pose at timestamp $k$%
\footnote{In this paper, we use the diacritics $\tilde{\cdot}$ and $\breve{\cdot}$ to represent the poses from SLAM and G2S prediction, respectively.}. 
Using $\tilde{\mathbf{T}}_{k}$ and the paired images $\{I^g_k, I^s_k\}$, we predict $\{\breve{x}_{k}, \breve{y}_{k}, \breve{\theta}_k\}$, indicating the G2S translation shifts (longitudinal and latitudinal) and the azimuth change w.r.t. $\tilde{\mathbf{T}}_{k}$%
\footnote{Here, we assume $\{I^g_0, I^s_0\}$ are aligned and set $\{x_{0}, y_{0}, \theta_0\}$ as $\{0, 0, 0\}$ for the first frame. }. 
Let $\mathbf{T}_{k} = [\mathbf{R}_{k}, \mathbf{t}_{k}; \mathbf{0}, 1]$ denote the updated poses, in the ideal case without noise, we have:
\begin{equation} \label{eq_fine}
\breve{\mathbf{R}}_{k}^\top = \mathbf{R}^\top_{k} \cdot \tilde{\mathbf{R}}_{k}, ~~
\breve{\mathbf{t}}_{k} = \tilde{\mathbf{R}}_{k}^\top \cdot (\mathbf{t}_{k} - \tilde{\mathbf{t}}_{k})
\end{equation}
where 
$\breve{\mathbf{T}}_{k} = [\breve{\mathbf{R}}_{k}, \breve{\mathbf{t}}_{k}; \mathbf{0}, 1]$ represents the G2S pose, $\breve{\mathbf{R}}_{k} =  \exp([\breve{\mathbf{r}}_{k}]^{\wedge})$, $\breve{\mathbf{r}}_{k} = [0, 0, \breve{\theta}_k]^\top$, 
$\breve{\mathbf{t}}_{k} = [\breve{x}_{k}, \breve{y}_{k}, 0]^\top$. Here, we do not update the translation along $z$-axis and the orientation around $x$,  $y$-axis%
\footnote{Owing to the (approximated) parallel projection of satellite images, the G2S registration focuses on a 3-DoF pose estimation.}. 


\section{Methodology}
This section introduces the details of the proposed framework. 
It consists of three main parts: deep learning based G2S registration, valid G2S pose selection, and the G2S-SLAM fusion. The main processes are outlined in Fig.~\ref{fig02_method}.


\subsection{G2S Registration} \label{subsec_g2sReg}
The G2S registration is based on BoostG2SLoc \cite{shi2023boosting}. 
From $I^g$, a U-Net architecture is adopted to extract the deep features $F^g$ which are then projected to the $I^s$ domain to synthesize an overhead view feature map $F^{g2s}$. Similarly, the deep features $F^s$ are extracted from $I^s$. The deep features $F^{g2s}$ and $F^s$ enable the network to implicitly learn the correspondence between $I^g$ and $I^s$ for pose prediction.

The camera poses are estimated in two steps. First, based on the deep feature $F_\text{init}^{g2s}$ (synthesized using the input pose) and $F^s$, a neural pose optimizer calculates the azimuth change $\breve{\theta}$ w.r.t. the input orientation. Second, the translation shift is calculated using an uncertainty-guided spatial correlation method. To be more specific, using the predicted $\breve{\theta}$, an overhead view feature map $F_{\breve{\theta}}^{g2s}$ is re-synthesized and used as a sliding window to compute its spatial correlation with $F^s$. A deep uncertainty map is adopted to exclude the infeasible vehicle locations, e.g., building roof and tree areas. 


\subsection{Valid G2S Pose Selection}
Since the satellite image does not provide 3D information, the valid cross-view features are mainly extracted from the ground, e.g., road marks, lanes, etc. When a vehicle moves forward, the feature changes along the longitudinal direction can be small. 
In other words, the translation changes on the input signals can be absorbed by the aggregation layers, and hence the predicted translation (especially along the longitudinal axis) can be noisy. To tackle this issue, this paper designs a coarse-to-fine method to remove the false G2S predictions by utilizing the SLAM information.

\subsubsection{Spatial Bound Check} \label{sec_sb}
BoostG2SLoc searches the G2S translation via a fixed-range spatial correlation. On the one hand, a large search range is needed such that the network is able to handle the long-term SLAM drift when the input pose is far from its ground truth. On the other hand, the large search range increases the chance of involving false deep features similar to the true one on $F_\theta^{g2s}$, which results in false G2S predictions. These false predictions need to be removed before combining with the SLAM measurements. 
To achieve this, we calculate a spatial bound proportional to the 3-$\sigma$ bound of the estimated trajectory covariance%
\footnote{
The trajectory estimation is illustrated in Sec. \ref{sec_online}. The covariance is from the information matrix at the solution of \eqref{eq_loss}. To be more specific, we calculate the covariance matrix using the inverse of the Hessian matrix. $\mathbf{\Phi}_k$ is then directly obtained from the covariance matrix.  }. 
%
When the input pose has low uncertainty, it is assumed to be close to the ground truth; therefore, a large G2S translation would be regarded as an incorrect prediction.
%
At timestamp $k$, suppose $\mathbf{\Phi}_k \in \mathbb{R}^{2\times 2}$ is the estimated covariance of the $x$-$y$ translation, the spatial bound is calculated based on \cite{zhang2018comparison}:
\begin{small}
\begin{equation} \label{eq_bound}
\mathbf{b}_k(\alpha) = \frac{3}{n}  
\begin{bmatrix}
\cos(\Theta(\mathbf{R}_k)) & -\sin(\Theta(\mathbf{R}_k)) \\
\sin(\Theta(\mathbf{R}_k)) & \cos(\Theta(\mathbf{R}_k)) \\
\end{bmatrix}
 \mathbf{\Phi}_k^{1/2} 
\begin{bmatrix}
\cos(\alpha) \\ \sin(\alpha)
\end{bmatrix}
\end{equation}
\end{small}
where $\Theta(\cdot)$ returns the azimuth angle from a rotation matrix, $\alpha \in [0, 2\pi)$, $n$ is the scale factor of the bound. Since the estimate by SLAM is usually accurate at the beginning, we assume the spatial bound $\mathbf{b}_1$ is within a fixed range $r$, and calculate a scale factor using $n = \text{mean}(\lambda_1, \lambda_2) / r$
%
where $\lambda_1, \lambda_2$ are the two eigenvalues of $\mathbf{\Phi}_1^{1/2}$. Let $\mathcal{B}_k$ denote the 2D space within the spatial bound, we have $\breve{\mathbf{t}}_{k} \in \mathcal{B}_k$ meaning the predicted translation is valid w.r.t. the spatial bound check.

\begin{algorithm}[t]  
	\caption{Iterative G2S-SLAM Fusion}
	\label{agm_fine}
	\LinesNumbered 
	\KwIn{SLAM poses, ground and satellite images.}
	\KwOut{Estimated vehicle trajectory.}
	Initialize poses $\mathbf{T}_k^0 = \tilde{\mathbf{T}}_k$; \\
	Initialize spatial bound $\mathcal{B}_k^0$;  \\
	Initialize G2S prediction for the first frame: $\breve{\mathbf{T}}_0^0 = \mathbf{I}_4$; \\
	Calculate visual odometry weights by Sec. \ref{sec_vow}; \\
	\For{all other poses} {
		\textbf{Step 1}: G2S Pose Prediction: \\
		Calculate G2S pose $\breve{\mathbf{T}}_k^t$ using the trajectory pose $\mathbf{T}_k^t$ and the images $\{I^g_k, I^s_k\}$; \Comment{{\textcolor{magenta}{`$t$' denotes the latest updated trajectory. $t=0$ at the beginning.}}} \\
		\textbf{Step 2}: Check validity: \\
		Calculate spatial bound $\mathcal{B}_{k-1}^t$ and $\mathcal{B}_{k}^t$; \\
		\If{$\breve{\mathbf{t}}^t_{k-1} \in \mathcal{B}^t_{k-1}$ \& $\breve{\mathbf{t}}^t_{k} \in \mathcal{B}^t_k$} {
			Calculate relative pose $\breve{\mathbf{T}}^t_{k-1,k}$ and $\mathbf{T}^t_{k-1,k}$;\\
			Visual odometry consistency check \eqref{eq_voc}; \\
		}
		\textbf{Step 3}: Trajectory Refinement: \\
		\If{$\breve{\mathbf{T}}_k^t$ is selected} {
			Solve the nonlinear least squares problem \eqref{eq_loss}; \\ 
			Update the spatial bound $\mathcal{B}^{t+1}_{t+1}, \cdots$; \Comment{{\textcolor{magenta}{For checking the rest G2S predictions.}}} \\
			Update all selected predictions $\mathcal{C}_r^{t+1}$, $\mathcal{C}_t^{t+1}$.
			\Comment{{\textcolor{magenta}{Making the selected predictions w.r.t. the latest trajectory for the next refinement.}}}
		}
	}
\end{algorithm}

\subsubsection{Visual Odometry Consistency} \label{sec_voc}
Although the spatial bound check can remove false G2S predictions inconsistent with the estimated uncertainty, it can not handle the case where false predictions are within the spatial bound%
\footnote{This can happen either because of the inaccuracy of the estimated uncertainty or because of multiple false deep features within a spatial bound.}. 
To tackle this, we further check the visual odometry consistency between the G2S predictions and the input poses, according to that the SLAM estimation is usually with acceptable accuracy in the short term. 
%
%
Based on \eqref{eq_fine}, the relative pose by cross-view is $\breve{\mathbf{T}}_{k-1,k}=\breve{\mathbf{T}}_{k-1}^{-1} \mathbf{T}_{k-1,k} \breve{\mathbf{T}}_{k}$, where $\mathbf{T}_{k-1,k}=\mathbf{T}_{k-1}^{-1}\mathbf{T}_k$ is the relative pose from the input trajectory. If $\breve{\mathbf{t}}_{k-1} \in \mathcal{B}_{k-1}$ and $\breve{\mathbf{t}}_{k} \in \mathcal{B}_k$, we respectively check if the rotation difference (azimuth angle) and the translation difference are smaller than a threshold. 
Let $\mathcal{C}_r$ and $\mathcal{C}_t$ denote the selected G2S orientation and translation, we have
\begin{equation} \label{eq_voc}
\begin{aligned}
\mathcal{C}_r &= \{\breve{\mathbf{R}}_k:~~ |\Theta(\breve{\mathbf{R}}_{k-1,k} \cdot \mathbf{R}_{k-1,k}^{\top})|<\text{th}_{\theta} \} \\
\mathcal{C}_t &= \{\breve{\mathbf{t}}_k:~~ |\mathbf{e}_i^{\top} \cdot (\breve{\mathbf{t}}_{k-1,k} - \mathbf{t}_{k-1,k})| <\text{th}_{t} \}
\end{aligned}	
\end{equation}
where $\mathbf{e}_1 = [1,0,0]^{\top}$ and $\mathbf{e}_2 = [0,1,0]^{\top}$ are for checking the translation consistency along $x$ and $y$ axis, respectively.

\subsection{Scaled Pose Graph based G2S-SLAM Fusion}
\subsubsection{Objective Function}
The vehicle trajectory is estimated by fusing the SLAM poses and the selected G2S predictions. 
For vSLAM, since solving a global Bundle Adjustment can be very expensive owing to the large number of feature points, a pose-graph is typically used for trajectory refinement, e.g., the loop closure refinement \cite{strasdat2010scale}. In this paper, we present a similar idea for G2S SLAM fusion, 
by solving a scaled pose graph problem. 

Suppose $\{\tilde{\mathbf{R}}_{i,j}, \tilde{\mathbf{t}}_{i,j}\}$ denotes a relative pose of visual odometry or loop closure from SLAM. $\tilde{\mathcal{V}}_r$ and $\tilde{\mathcal{V}}_t$ are the set of all SLAM relative rotations and translations, respectively. The vehicle poses $\{\mathbf{R}_{k}, \mathbf{t}_{k}\}$ are calculated by
\begin{small}
\begin{equation} \label{eq_loss}
\begin{aligned}
	\argmin_{ \{\cdots, \mathbf{R}_{k}, \mathbf{t}_{k}, s_{k}, \cdots\} } 
&\sum_{\tilde{\mathbf{R}}_{i,j} \in \tilde{\mathcal{V}}_r}
\| \tilde{w}_{i,j} 
[\log(   \tilde{\mathbf{R}}_{i,j} \cdot  \mathbf{R}_{j}^{\top} \cdot \mathbf{R}_{i} )]^{\vee}
\|^2_{\tilde{\mathbf{\Sigma}}_r}
\\  + &
\sum_{\tilde{\mathbf{t}}_{i,j} \in \tilde{\mathcal{V}}_t} 
\|\tilde{w}_{i,j} (
\tilde{\mathbf{t}}_{i,j} - s_j \mathbf{R}_{i}^{\top} (\mathbf{t}_{j} - \mathbf{t}_{i})
)\|^2_{\tilde{\mathbf{\Sigma}}_t}
\\  + &
\sum_{\breve{\mathbf{R}}_{l} \in \breve{\mathcal{C}}_r} 
\|
[\log(\mathbf{R}_{l}^{\top} \cdot \tilde{\mathbf{R}}_{l} \cdot \breve{\mathbf{R}}_{l}^{\top})]^{\vee}
\|^2_{\breve{\mathbf{\Sigma}}_r}
\\  + &
\sum_{\breve{\mathbf{t}}_{l} \in \breve{\mathcal{C}}_t} \rho (
\|
(
\breve{\mathbf{t}}_{l} - \tilde{\mathbf{R}}_{l}^{\top} (\mathbf{t}_{l} - \tilde{\mathbf{t}}_{l})
)\|^2_{\breve{\mathbf{\Sigma}}_t}
)
\\  + &
\sum_{k=1}^{K} 
\|s_k - s_{k-1}
\|^2_{\sigma_s}
\end{aligned}
\end{equation}
\end{small}
where $\log: \text{SO}(3) \rightarrow \mathfrak{so}(3)$ is the logarithm map and $[\cdot]^{\vee}$ returns the vector elements from a skew-symmetric matrix. 
$\tilde{\mathbf{\Sigma}}_r = \tilde{\sigma}_r\mathbf{I}_3$, 
$\tilde{\mathbf{\Sigma}}_t = \tilde{\sigma}_t\mathbf{I}_3$, 
$\breve{\mathbf{\Sigma}}_r = \breve{\sigma}_r\mathbf{I}_3$, 
$\breve{\mathbf{\Sigma}}_t = \text{diag}(\breve{\sigma}^x_t, \breve{\sigma}^y_t, 0)$, 
and $\sigma_s$ are the hyper-parameters to balance the terms. 

In \eqref{eq_loss}, the first two terms represent the measurements by SLAM, where we also estimate a scale $s_k$ for each pose to reflect the trajectory drift of vSLAM%
\footnote{A similar idea is shown in \cite{strasdat2010scale} for monocular SLAM. We find that even for stereo SLAM where the scale is observable, recovering the scale for each pose is helpful to reduce the trajectory error in our framework.}.
Since the SLAM drift is usually smooth, we restrict the scale change among the neighbouring frames using the fifth term.

The third and the fourth terms represent the measurements from G2S predictions based on \eqref{eq_fine}. 
For the G2S translation term, we assign $\breve{\sigma}^x_t<\breve{\sigma}^y_t$ since the latitudinal predictions are usually with higher accuracy than the longitudinal predictions. A Huber kernel is used to bring more robustness:
\begin{equation} \label{eq_huber}
\rho(x) = 
\left\{
\begin{matrix}
x^2/2           & |x|<c \\
c\cdot(|x| - c/2)    & |x|\geqslant c \\
\end{matrix}
\right.
\end{equation}

\subsubsection{Visual Odometry Weights} \label{sec_vow}
We follow \cite{chen2021anchor} to calculate the visual odometry weights. Suppose $N_{i,j}$ denotes the number of covisible features between two frames $i$ and $j$, $\tilde{w}_{i,j} = \sqrt{N_{i,j}} / \tilde{n}_{i,j}$, where the factor $\tilde{n}_{i,j} = \text{mean}(\cdots, \sqrt{N_{i,j}}, \cdots)$ is to make the weights stable among different dataset.

\subsubsection{Optimization} \label{sec_opti}
The non-linear least squares problem \eqref{eq_loss} can be solved iteratively using the Gauss-Newton method, where the robust kernel can be represented by reweighting the measurement term \cite{chebrolu2021adaptive}. Suppose $\mathbf{x}$ denotes the concatenation of all state variables, $\mathbf{W}$ denotes the stacked weights. In each iteration, the solver linearises the problem at $\mathbf{x}$ by $\mathbf{J}(\mathbf{x})$ (the stacked Jacobian matrix), and calculates the step change to update $\mathbf{x}$: 
\begin{equation}
	\Delta \mathbf{x} = -(\mathbf{J}(\mathbf{x})^{\top} \cdot \mathbf{W} \cdot \mathbf{J}(\mathbf{x}))^{-1} \cdot (\mathbf{J}(\mathbf{x})^{\top} \cdot \mathbf{W} \cdot \mathbf{f}(\mathbf{x}))
\end{equation}
where $\mathbf{f}(\mathbf{x})$ is the stacked residual. At the solution point $\mathbf{x}^{*}$, we can calculate the estimated covariance matrix $(\mathbf{J}(\mathbf{x}^{*})^{\top} \cdot \mathbf{W} \cdot \mathbf{J}(\mathbf{x}^{*}))^{-1}$ for the spatial bound check in Sec. \ref{sec_sb}.


\subsubsection{Iterative Fusion Pipeline} \label{sec_online}
The accuracy of G2S prediction decreases when the trajectory drift becomes out of the search range by BoostG2SLoc. To tackle this issue, we provide an iterative trajectory fusion pipeline. At the beginning, we initialize the vehicle poses $\mathbf{T}_k^0$ using SLAM trajectory. The spatial bound $\mathcal{B}_k^0$ is also initialized using the covariance matrix%
\footnote{Using the SLAM poses, we calculate the Hessian matrix of \eqref{eq_loss} without the cross-view measurements}.
Suppose at timestamp $t$, the predicted G2S pose is selected. We update the trajectory using all the current selected G2S measurements and the results from SLAM by solving \eqref{eq_loss}. Having the new trajectory $\mathbf{T}_k^{t+1}$ ($k=\{1,\cdots, K\}$), we can calculate the spatial bounds $\mathcal{B}^{t+1}_{t+1}, \cdots, \mathcal{B}^{t+1}_{K}$ (only the spatial bounds for the following poses are needed for the next spatial bound check). We also update the G2S measurements $\mathcal{C}_r^{t+1}$ and $\mathcal{C}_t^{t+1}$ (all selected G2S measurements are used for the next update). More details are shown in Algorithm \ref{agm_fine}.

\begin{table*}[t]  
\setlength{\abovecaptionskip}{0pt}
    \setlength{\belowcaptionskip}{0pt}
	\caption{ Accuracy Comparison with SLAM Result using KITTI Dataset}
	\label{tab_compare_slam}
	\begin{center}
			\begin{tabular}{c| c c c c c c  |  c c c c c c}
				\hline \hline
				\multirow{2}{*}{Sequence} & \multicolumn{6}{c}{$\mathbf{S}$ by Trajectory Origin} & \multicolumn{6}{c}{$\mathbf{S}$ by Multiple Ground Truth} \\
				\cline{2-13}
				& $\theta^\dag~\downarrow$ & $\theta^\S~\downarrow$ & $\theta^\pounds \%~\uparrow$   
				& $\mathbf{t}^\dag~\downarrow$ & $\mathbf{t}^\S~\downarrow$ & $\mathbf{t}^\pounds \% ~\uparrow$ 
				& $\theta^\dag~\downarrow$ & $\theta^\S~\downarrow$ & $\theta^\pounds \%~\uparrow$   
				& $\mathbf{t}^\dag~\downarrow$ & $\mathbf{t}^\S~\downarrow$ & $\mathbf{t}^\pounds \%~\uparrow$ \\
				\hline
				00  & 0.731 & 0.491 & 32.7 \%   & 4.144  & 0.946  & 77.2 \% & 0.545  & 0.545 & -0.2\%  & 1.081 & 1.306 & -20.8\% \\
				01  & 2.644 & 0.726 & 72.6 \%   & 31.978 & 1.516  & 95.3\%  & 1.733  & 0.717 & 58.6\%  & 15.00 & 1.529 &  89.8\% \\
				02  & 1.001 & 0.211 & 78.9 \%   & 5.650  & 0.823  & 85.4\%  & 0.547  & 0.214 & 61.0\%  & 3.661 & 0.802 &  78.1\% \\
				04  & 0.037 & 0.188 & -409.8\%  & 0.625  & 0.363  & 41.9\%  & 0.413  & 0.088 & 78.6\%  & 0.193 & 0.344 & -78.8\% \\
				05  & 0.304 & 0.231 & 23.9\%    & 1.292  & 0.543  & 58.0\%  & 0.257  & 0.288 & -12.1\% & 0.922 & 0.435 &  52.8\% \\
				06  & 0.832 & 0.430 & 48.3\%    & 2.657  & 1.311  & 50.6\%  & 0.426  & 0.367 & 14.0\%  & 0.912 & 0.768 &  15.7\% \\
				07  & 0.291 & 0.203 & 30.4\%    & 0.640  & 0.512  & 20.0\%  & 0.296  & 0.278 & 6.2\%   & 0.419 & 0.355 &  15.3\% \\
				08  & 1.990 & 0.466 & 76.6\%    & 7.460  & 1.168  & 84.4\%  & 1.038  & 0.523 & 49.6\%  & 4.323 & 2.073  & 52.1\% \\
				09  & 0.949 & 0.205 & 78.4\%    & 3.080  & 1.553  & 49.6\%  & 0.779  & 0.286 & 63.3\%  & 1.903 & 1.254  & 34.1\% \\
				10  & 0.784 & 0.146 & 81.3\%    & 3.538  & 0.646  & 81.7\%  & 0.461  & 0.190 & 58.7\%  & 0.906 & 0.509  & 43.8\% \\
				\hline
				Avg.  & 0.956 & {\bf 0.330} & 65.5\%  & 6.106  & {\bf 0.938}  & 84.6\%  & 0.650  & {\bf 0.350} & 46.2\%  & 2.932 & {\bf 0.938}  & 68.0\% \\
				\hline \hline
			\end{tabular}
		\begin{tablenotes}
			\item[1] $\theta$: RMSE of absolute azimuth rotation (unit: $^\circ$); $\mathbf{t}$ RMSE of absolute 2D translation (unit: m);
			\item[2] $\dag$: The stereo SLAM result using \cite{campos2021orb}; $\S$: the result by the proposed method; 
			\item[3] $\pounds$: the accuracy improvement using $\frac{\text{SLAM error} - \text{our method error}}{\text{SLAM error}}$;
			\item[4] `$\downarrow$': smaller error represents higher accuracy; `$\uparrow$': higher percentage represents larger improvement. 
		\end{tablenotes}
	\end{center}
\end{table*}

\section{Experiments and Results}
In this section, the evaluations using real datasets are presented. 
We first introduce the details of data preparation, network referencing, and hyper-parameter setting. The experiment results are then presented followed by an ablation study showing the effectiveness of each module in the framework. For all experiments, we use the stereo SLAM trajectories by ORB-SLAM3 \cite{campos2021orb} and G2S predictions by BoostG2SLoc \cite{shi2023boosting}. 
However, we should note that our method is not limited to any specific SLAM or cross-view registration framework. 

\subsection{Experimental Setup}

\subsubsection{Dataset Preparation}
The framework is evaluated using two publicly available autonomous driving datasets KITTI \cite{geiger2013vision} and FordAV \cite{agarwal2020ford}. 
The satellite images are collected from Google Map \cite{gm}. Each satellite image covers a region around $100m \times 100m$ with a resolution of $512 \times 512$. The ground-view images are re-sized to $256\times1024$ for network referencing. More details are shown in \cite{shi2022beyond}. 

\subsubsection{G2S Registration}
The G2S registration is performed by loading the weights pre-trained on the KITTI and FordAV from \cite{shi2023boosting}%
\footnote{All experiments are conducted on a desktop with the Intel(R) Core(TM) i7-13700KF CPU and the GeForce RTX 3090 GPU.}.
For all experiments, we set the G2S search range for rotation as $\pm 10^{\circ}$, and the location search range as $20m\times 20m$. We use the same-area G2S model for KITTI and FordAV, respectively.






\subsubsection{Parameter Setup}
We set $r=0.01$, $\text{th}_{\theta} = 0.25$, $\text{th}_{t} = 0.5$ in \eqref{eq_voc}; $\tilde{\sigma}_r=0.85^2$, $\tilde{\sigma}_t=0.9^2$, $\breve{\sigma}_r=1$, $\breve{\sigma}^x_t=0.003^2$, $\breve{\sigma}^y_t=0.005^2$, $\sigma_s=10^2$ in \eqref{eq_loss} and $c=1$ in \eqref{eq_huber} on KITTI. 
For the experiments on FordAV, we set $r=0.2$, $\text{th}_{\theta} = 0.5$, $\breve{\sigma}^x_t=0.001^2$, $\sigma_s=9^2$, $c=6$, and keep the rest hyper-parameters unchanged.
The parameters for different trajectories in each dataset are the same. 


\subsection{Evaluation Metrics}
For autonomous driving scenarios, the localization accuracy on the $x$-$y$ plane is of more concern. This paper adopts two metrics to evaluate the trajectory 2D accuracy.

The absolute mean, median, and root-mean-square error (RMSE) of rotation and translation are used to measure the error between the estimated trajectory and the ground truth%
\footnote{To better reflect the global localization accuracy, we use the absolute error rather than the relative translation/rotation error for evaluation.}:
\begin{equation} \label{eq_evaluation}
\delta \mathbf{T} = \check{\mathbf{T}}^{-1} \mathbf{S} \mathbf{T}
\end{equation}
where $\mathbf{T}$ and $\check{\mathbf{T}}$ are the estimated poses and the ground truth.  $\mathbf{S}$ is the rigid-body transformation mapping the estimated trajectory to the ground truth \cite{sturm2012benchmark}. $\mathbf{S}$ is obtained either using the ground truth pose at the trajectory origin, or using multiple ground truth poses by solving a least-squares problem \cite{horn1988closed}. 

Following \cite{shi2023boosting} and \cite{shi2022beyond}, we also present the translation error along the longitudinal and lateral directions, and the rotation error of azimuth orientation. Specifically, we show the percentage of the correctly estimated translation and rotation within $1$ meter and $1^{\circ}$.

\subsection{Assessment on G2S Pose Selection}
\begin{figure}[t]  
\centering
\includegraphics[width=1.0\linewidth]{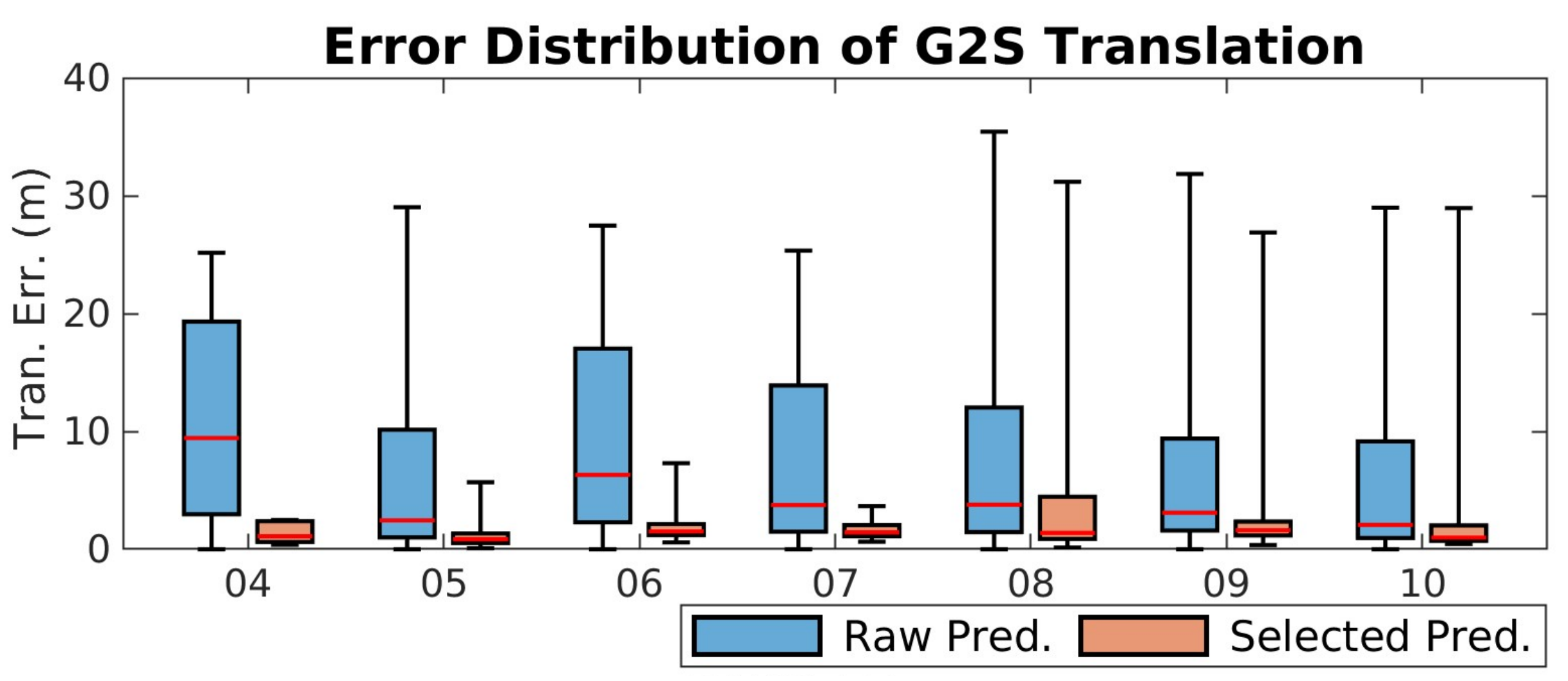}    
\caption{Error distribution (unit m) of the raw G2S predictions and that of the selected predictions. Here, we use seven sequences (`04'-`10') from KITTI Odometry Benchmark for evaluation. We do not update the trajectory to avoid the effect from other modules.
 }
\label{fig_inlier_cvpose}
\end{figure}

The valid cross-view selection is important to increase the accuracy of trajectory estimation. To evaluate the proposed G2S pose selection module, we report the error distribution of the raw predictions and the selected G2S poses in Fig. \ref{fig_inlier_cvpose}. Overall, the selected poses are with smaller errors, indicating that most false G2S predictions are effectively removed.

\begin{figure}[!thpb]  
		\centering
  \setlength{\abovecaptionskip}{0pt}
    \setlength{\belowcaptionskip}{0pt}
		\includegraphics[width=1.0\linewidth]{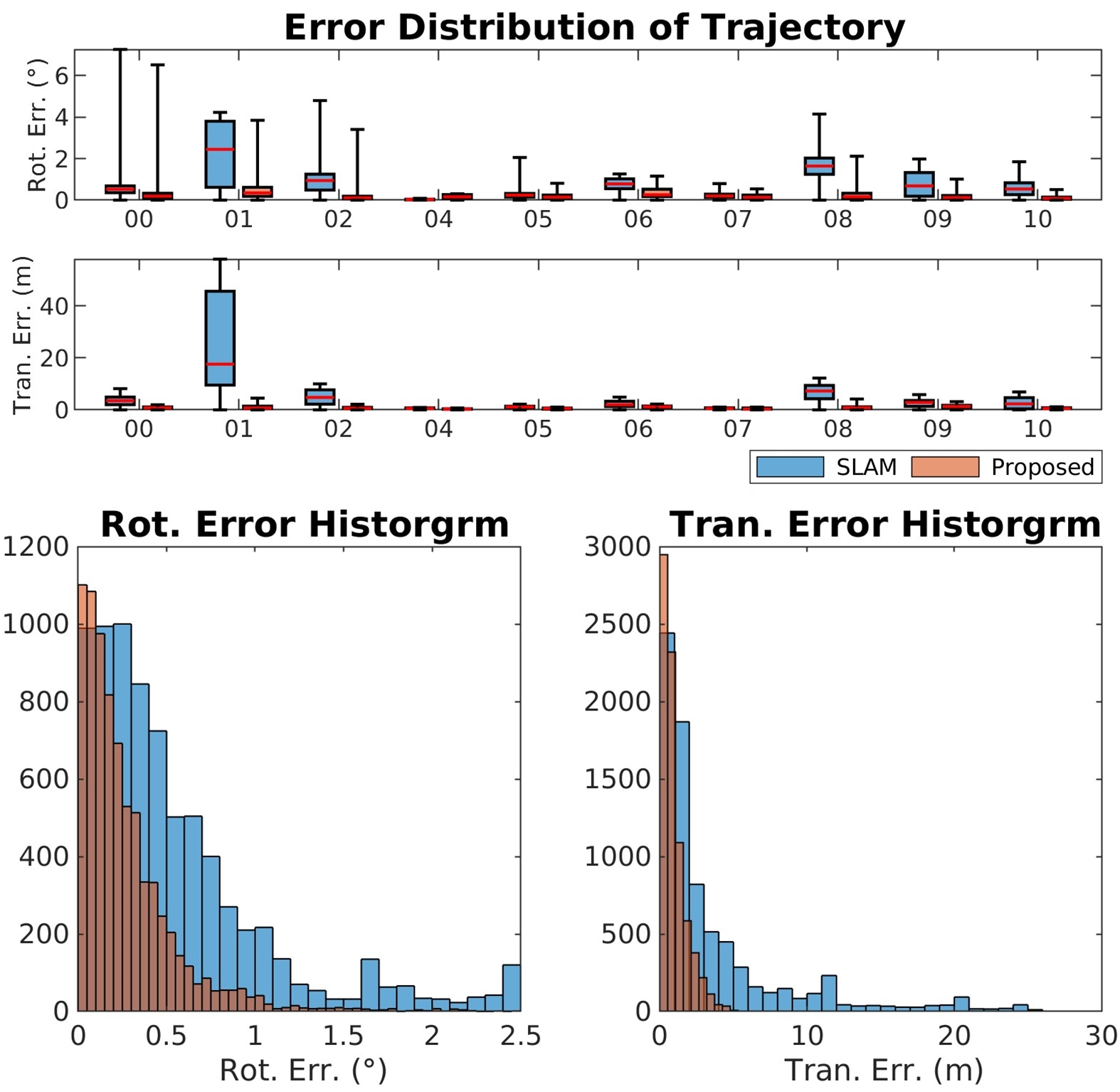}    
		\caption{ A comparison of localization error distribution. The RMSE is shown in Table \ref{tab_compare_slam}. The 1st-2nd rows are the rotation (unit $^\circ$) and translation (unit m) error distribution of each sequence, where $\mathbf{S}$ is by trajectory origin. The 3rd row reports the histograms of all rotation and translation errors, where $\mathbf{S}$ is by multiple ground truth. Overall, the error by the proposed framework is lower and more concentrated.
		}
		\label{fig_slam_distribution}
\end{figure}

\begin{figure*}[t]  
 \setlength{\abovecaptionskip}{0pt}
    \setlength{\belowcaptionskip}{0pt}
		\centering
		\includegraphics[width=0.95\linewidth]{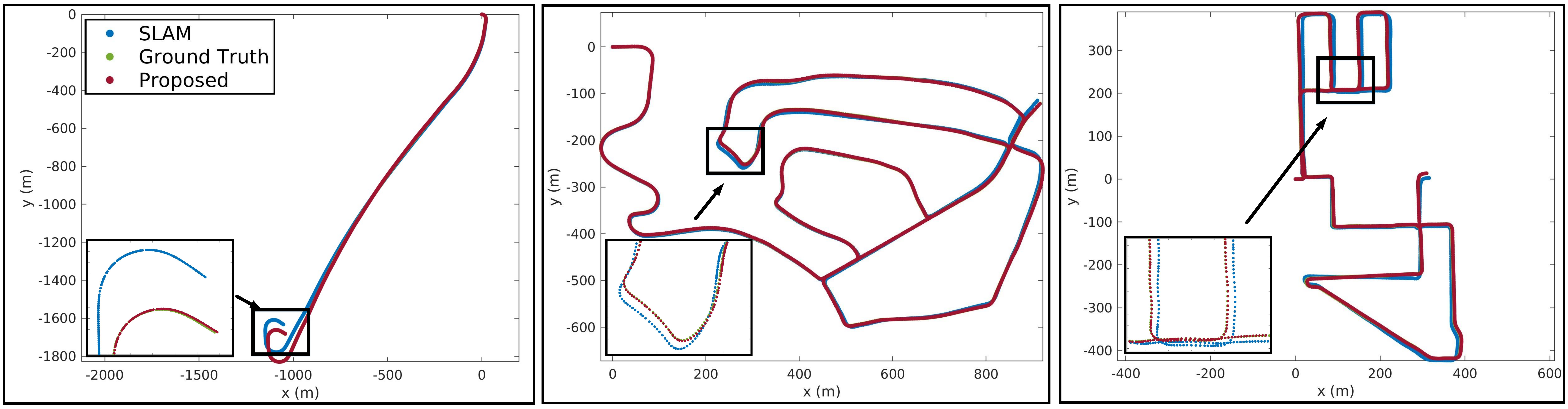}    
		\caption{ Examples of the estimated trajectories on KITTI. The first figure shows a scenario without loop closure. 
  For the second and the third figures, the trajectories estimated by SLAM are with loop closure.
  For all results, our estimated trajectories (red) are very close to the ground truth (green). 
		}
		\label{fig_kitti_rst}
	\end{figure*}	



\subsection{Assessment on Trajectory Estimation}
Table \ref{tab_compare_slam} reports a comparison of the estimated rotation and translation RMSE using the stereo SLAM and the proposed framework. To fully reflect the accuracy, 
we calculate $\mathbf{S}$ in \eqref{eq_evaluation} using the ground truth at trajectory origin (better reflecting the true estimate as minimum ground truth information is involved) and multiple ground truth poses (common among literature)%
\footnote{In this paper, we present the result using $\mathbf{S}$ by trajectory origin, unless otherwise noted.}.
Overall, the proposed framework achieves higher accuracy for vehicle localization. On average, the translation error reduces $64\%$ (from $0.96^{\circ}$ to $0.34^{\circ}$) and the rotation error reduces $83\%$ (from $6m$ to $1m$). Fig. \ref{fig_slam_distribution} shows a comparison of the error distribution. Overall, the localization error by the proposed framework is lower and more concentrated, which illustrates the robustness of our method.

\begin{table}[!thpb]
\setlength{\abovecaptionskip}{0pt}
    \setlength{\belowcaptionskip}{0pt}
    \setlength{\tabcolsep}{5pt}
	\caption{ Accuracy Comparison with G2S Prediction}
	\label{tab_compare_cv}
	\begin{center}
			\begin{tabular}{c| c c | c c | c c }
				\hline \hline
				\multirow{2}{*}{Meth.} & \multicolumn{2}{c}{Azimuth ($^\circ$)} & \multicolumn{2}{c}{Longitudinal (m)} & \multicolumn{2}{c}{Lateral (m)} \\
				\cline{2-7} 
				& mean$\downarrow$  & 1$^{\circ}(\%)\uparrow$ 
				& mean$\downarrow$ & 1m $(\%)\uparrow$ 
				& mean$\downarrow$ & 1m $(\%)\uparrow$  \\
				\hline
				\cite{shi2023boosting}  
				& 0.163 & 99.9\%  
				& 7.651 & 20.4\% 
				& 0.746 & 79.6\%  \\
				Ours  
				& 0.231 & 98.0\%  
				& 0.544 & 84.1\% 
				& 0.485 & 89.9\%  \\
				\hline \hline
			\end{tabular}
		\begin{tablenotes}
			\item[1] We report the mean and the percentage of the estimated results larger than the threshold. Results are averaged across ten KITTI sequences. 
		\end{tablenotes}
	\end{center}
\end{table}

\begin{figure}[!thpb]  
\centering
\setlength{\abovecaptionskip}{0pt}
    \setlength{\belowcaptionskip}{0pt}
\includegraphics[width=1.0\linewidth]{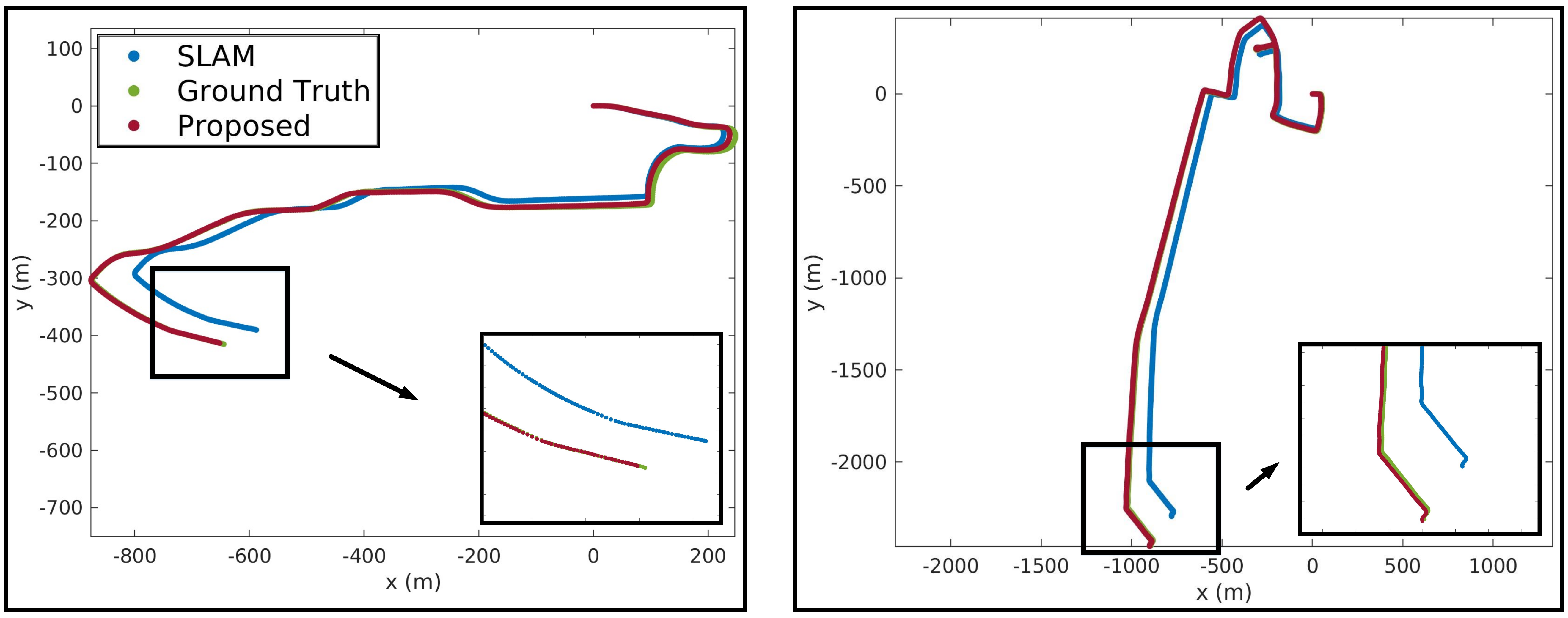}    
\caption{ 
Evaluation on FordAV. On average, the RMSE is \textcolor{myBlue}{\{$1.08^{\circ}$, $65.24$m\}} by  \cite{campos2021orb}, and \textcolor{myRed}{\{$0.80^{\circ}$, $7.17$m\}} by the proposed method. 
}
\label{fig_ford_rst}
\end{figure}

Fig. \ref{fig_kitti_rst} and Fig. \ref{fig_ford_rst} visually report the estimated trajectory on KITTI and FordAV datasets%
\footnote{
FordAV is more challenging than KITTI for visual localization, e.g., the rapid illumination changes between the consecutive images can cause tracking loss of the vSLAM system. Because of this, the evaluation is conducted using part of the trajectories without tracking loss from SLAM. 
}. 
We can see that the proposed framework improves the localization accuracy, especially for the scenario without loop closure (the 1st figure in Fig. \ref{fig_kitti_rst} and all figures in Fig. \ref{fig_ford_rst}) which is common in real applications for autonomous driving. 




Table \ref{tab_compare_cv} presents a comparison of the estimated trajectory with the G2S prediction in terms of the longitudinal, lateral, and azimuth error. We can see that the translation estimates (especially along the longitudinal direction) by the proposed methods are with higher accuracy, while the rotation error using both methods is small.

\subsection{Ablation Study}
We conduct an ablation study to evaluate the contribution of each module in the proposed framework. The translation errors by removing different configurations are present in Table. \ref{tab_ablation_trajectory}. We can see that the result using the full proposed framework achieves the best performance.

\begin{table}[!htbp]
\setlength{\abovecaptionskip}{0pt}
    \setlength{\belowcaptionskip}{0pt}
\caption{Ablation Study on Different Configurations}
\label{tab_ablation_trajectory}
\begin{center}
\begin{tabular}{c| c c c c c | c }
\hline \hline
& All G2S & SPB & VOC  & No $s$ & Non-Iter & Full \\
\hline
mean$\downarrow$   & 2.557 & 1.708 & 1.190 & 2.521 & 1.786 & 0.808 \\
median$\downarrow$ & 1.703 & 1.838 & 1.103 & 2.278 & 1.121 & 0.714 \\
RMSE$\downarrow$   & 3.367 & 1.880 & 1.352 & 2.953 & 2.288 & 0.938 \\
\hline \hline
\end{tabular}
\begin{tablenotes}
\item[1] 
All G2S: non-iterative G2S-SLAM fusion using all G2S poses; 
SPB: using only spatial bound to check G2S poses; 
VOC: using only visual odometry consistency to check G2S poses; 
No $s$: without scale estimate in \eqref{eq_loss}; 
Non-Iter: non-iterative G2S-SLAM fusion (updating the trajectory once using all selected G2S poses, rather than the pipeline in Sec. \ref{sec_online}); Full: results using all proposed modules.
\item[2]  Results are averaged across ten KITTI sequences. 
\end{tablenotes}
\end{center}
\end{table}



\section{Limitation}
Although achieving promising results, there are several limitations. First, the reliance on SLAM for G2S selection means that tracking loss within SLAM will impact the proposed method, particularly we find that some challenges causing tracking loss (e.g., illumination changes from the ground view images) also lead to inaccurate G2S predictions. Second, the proposed method requires more computational resources. Finally, satellite images are not accessible for certain environments, e.g., tunnels or indoor parking areas.


\section{Conclusion}
This paper proposes a framework for vehicle localization. The framework combines the stereo SLAM and the G2S cross-view registration to improve the camera localization accuracy. The G2S poses are predicted using a deep-learning based method, and their validities are checked using a coarse-to-fine method. The selected prediction is then fused with the SLAM poses by solving a scaled pose graph problem. Detailed validation using real experiments is conducted, and the results illustrate the localization accuracy as well as the potential value of this framework to be applied for autonomous driving. 

%
%
%
%
%

In the future, we plan to investigate a more tightly coupled fusion method by combining the 3D maps by SLAM. We also plan to investigate more advanced G2S methods. Our goal is to develop a SLAM-G2S-Fusion system for autonomous driving.




\balance
\bibliographystyle{./bibliography/IEEEtran}
\bibliography{./bibliography/IEEEabrv,./bibliography/mybibfile}

\end{document}